\documentclass[12pt]{article}

\usepackage{sbc-template}
\usepackage{graphicx,url}
\usepackage{sidecap}
\sidecaptionvpos{figure}{t}
\usepackage[utf8]{inputenc}  
\usepackage[numbers]{natbib}

\title{Fusarium Damaged Kernels Detection Using Transfer Learning on Deep Neural Network Architecture}

\author{Márcio Nicolau\inst{1,2}, Márcia Barrocas Moreira Pimentel\inst{1}, Casiane Salete Tibola\inst{1}\\ 
José Mauricio Cunha Fernandes\inst{1,2}, Willingthon Pavan\inst{2}}

\address{Brazilian Agricultural Research Corporation (Embrapa)\\
Passo Fundo -- Rio Grande do Sul State -- Brazil
\nextinstitute
Graduate Program in Applied Computing (PPGCA), \\
University of Passo Fundo (UPF)\\
Passo Fundo -- Rio Grande do Sul State -- Brazil
  \email{\{marcio.nicolau,marcia.pimentel,casiane.tibola,mauricio.fernandes\}@embrapa.br}
  \email{pavan@upf.br}
}

\begin{document} 

\maketitle

\begin{abstract}
The present work shows the application of transfer learning for a pre-trained deep neural network (DNN), using a small image dataset ($\approx$ 12,000) on a single workstation with enabled NVIDIA GPU card that takes up to 1 hour to complete the training task and archive an overall average accuracy of 94.7\%. The DNN presents a 20\% score of misclassification for an external test dataset. The accuracy of the proposed methodology is equivalent to ones using HSI methodology (81\%-91\%) used for the same task, but with the advantage of being independent on special equipment to classify wheat kernel for FHB symptoms. \\
\\{\small {\bf Keynotes.} Deep Learning, Fusarium Damaged Kernels, ImageNet, TensorFlow, Transfer Learning.}
\end{abstract}

\section{Introduction}
The wheat is the main source of nutrients to the world population, most of its production is converted into flour for human consumption. In Southern Brazil, where 90\% of the domestic wheat is produced, Fusarium head blight (FHB), a fungal disease, is a major concern. Apart from yield loss, the causal agent Fusarium graminearum may cause mycotoxin contamination of wheat products, creating health problems.

Therefore, to avoid potential health risks, Fusarium affected grains must be identified and segregated, before their processing, to avoid its incorporation into food for humans and animal feed.

Usually, the detection of Fusarium head blight (FHB) is carried out manually by human experts using a process that may be both lengthy and tiresome. Moreover, the effectiveness of this kind of detection may drop with factors such as fatigue, external distractions and optical illusions ~\cite{barbedo2015detecting}. Thus, improving the detection of Fusarium Head Blight (FHB) in wheat kernels has been a major goal, due to the health risks associated with the ingestion of the mycotoxin, mainly deoxynivalenol (DON).

Methods capable of performing this disease detection in an automatic way are highly demanded in the productive wheat chain, to segregate lots. Most automatic methods proposed to date rely on image processing to perform their tasks \cite{barbedo2015detecting,barbedo2017deoxynivalenol,maloney2014digital}.

\citet{barbedo2015detecting} used Hyperspectral imaging (HSI) for detecting Fusarium head blight (FHB) in wheat kernels using an algorithm. The outcome was a Fusarium index (FI), ranging from 0 to 1, that can be interpreted as the likelihood of the kernel to be damaged by FHB. According to the authors, hyperspectral imagery is currently not sensitive enough to estimate DON content directly. However, an indirect estimation from the Fusarium damaged kernels was successfully achieved, with an accuracy of approximately 91\% \cite{barbedo2015detecting}.

Other study investigated the use of hyperspectral imaging (HSI) for deoxynivalenol (DON) screening in wheat kernels. The developed algorithm achieved accuracies of 72\% and 81\% for the three- and two-class classification schemes, respectively. The results, although not accurate enough to provide conclusive screening, indicating that the algorithm could be used for initial screening to detect wheat batches that warrant further analysis regarding their DON content \cite{barbedo2017deoxynivalenol}.

\citet{min2015spectroscopic} presented a review about nondestructive detection of fungi and mycotoxins in grains, focusing on spectroscopic techniques and chemometrics. The spectroscopy has advantages over conventional methods including the rapidness and nondestructive nature of this approach. However, some limitations as expensive and complex setup equipment’s and low accuracy due to external interferents exist, which must be overcome before widespread adoption of these techniques.

The application of computer vision on digital images offers a high-throughput and non-invasive alternative to analytical and immunological methods. This paper presents an automated method to detect Fusarium Damaged Kernels, which uses the application of computer vision to digital images.

The main goal of this work is the use of machine learning algorithms and computer vision techniques to detect Fusarium Damaged Kernels in wheat, based on digital images.
\section{Material and Methods}
Digital images of Fusarium Head Blight symptomatic and non-symptomatic wheat kernels were available at the National Research Centre for Wheat (Embrapa Wheat), located in Passo Fundo, Rio Grande do Sul State, Brazil.

The images were obtained by recording a video of 06:25 minutes using an Olympus SP-810UZ digital camera with 36x optical zoom, 24mm wide-angle view, and 14-megapixel resolution, 720p HD video and Olympus Lens 36x Wide Optical Zoom ED 4.3-154.8mm 1:2.9-5.

All tasks run on an iMac workstation configured with 32GB of RAM DDR3 1600MHz, a 3.5GHz quad-core Intel Core i7 processor, and a NVIDIA GeForce GTX 780M GPU with 4GB of GDDR5 memory. The TensorFlow 1.0.1 built from source with CUDA Toolkit 8.0 and cuDNN v5.1 to enable GPU support. All scripts were developed using Python 2.7.
\subsection{Methodology} 
To classify digital images in predefined classes, we could use one of the several methods developed in recent two decades \cite{DNNSchmidhuber2015,DeepVisualGuo2016}. Methodologies to solve this kind of problem was developed both in Artificial Intelligence (AI), a research area in Computer Science, and in Statistics.

The main differences between Statistics and AI approach are the size of the task, for statistics point of view, algorithms and techniques are limited when then input size of image dataset are greater than tens of thousand pictures and for a large number of classification sets.

During the training stage of a system to classify images and objects based only on information content embedded in a single digital file, it is necessary that this system would detect all possible contexts where the object could occur. Small differences in color, luminosity, angle and other could be misinterpreted by the proposed system and results in the wrong or low-level prediction classification.

The key advantage of using AI strategy in the image classification task are related to knowing how to combine layers and manage the relationship between levels of information, without needing any strong assumption related to the type of dependency or relational structure among input information.

In some cases, the improvements in the accuracy and precision gain are archive using these fine tune setting, but the most of the effort is made only with the input information, in other words, the processes are designed to take the most of the self-learning way.  

\subsection{Deep Convolution Neural Network (DNN)}
The most used architecture of the neural network for image classification task is called convolutional neural network (CNN). The convolution operation or, sometimes called convolution layer is related to the operation to process or respond to ``stimuli''  in a limited region known as the receptive field.

The receptive field from each neuron contains a partial overlap of information from input layer (raw image) and, in this way, the preprocessing or further operations occur with a minimal amount of effort. Other advantages of this technique are related to the possibility of using distributed algorithms (even GPU versions) to calculate, filters and processing small pieces of information, one each time and aggregated the results when necessary \cite{krizhevsky2012imagenet}.

\begin{figure}[htb]
\centering
\caption{An example of relationship between layers in a CNN with convolution, max pooling and dense neuron connections annotated on the illustration.}\label{fig:figCNN}
\includegraphics[width=1.0\textwidth]{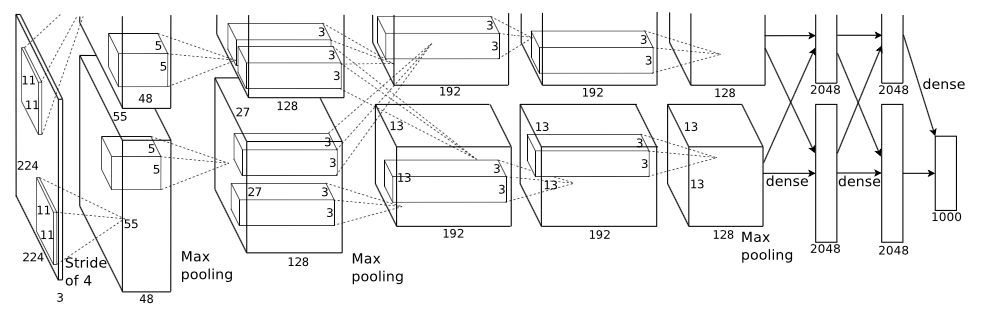}
\begin{center}Source: \cite[Figure 2]{krizhevsky2012imagenet}\end{center}
\end{figure}

The deep portion of CNN come from the stacking or combination of several layers where the output of preceding layer is used as an input for the next one. The most common layers employed in DNN are convolution, ReLU (Rectified Linear Units), tanh (Hyperbolic Tangent Function), max pooling, average pooling, fully connected, concat, dropout and softmax. To better understand this relationship see a CNN example in Figure~\ref{fig:figCNN}.

%
\subsection{Transfer Learning}
\begin{figure}[htb]
\centering
\caption{An illustration of a representation of a Deep Convolutional Neural Network, trained on top of ImageNet with detailed information about Feature Extraction part and Classification part. Each small box represents one of the layer types in the Inception Network: Convolution AvgPool, MaxPool, Concat, Dropout, Fully Connected and Softmax.}\label{fig:featureExtraction}
\includegraphics[width=1.0\textwidth]{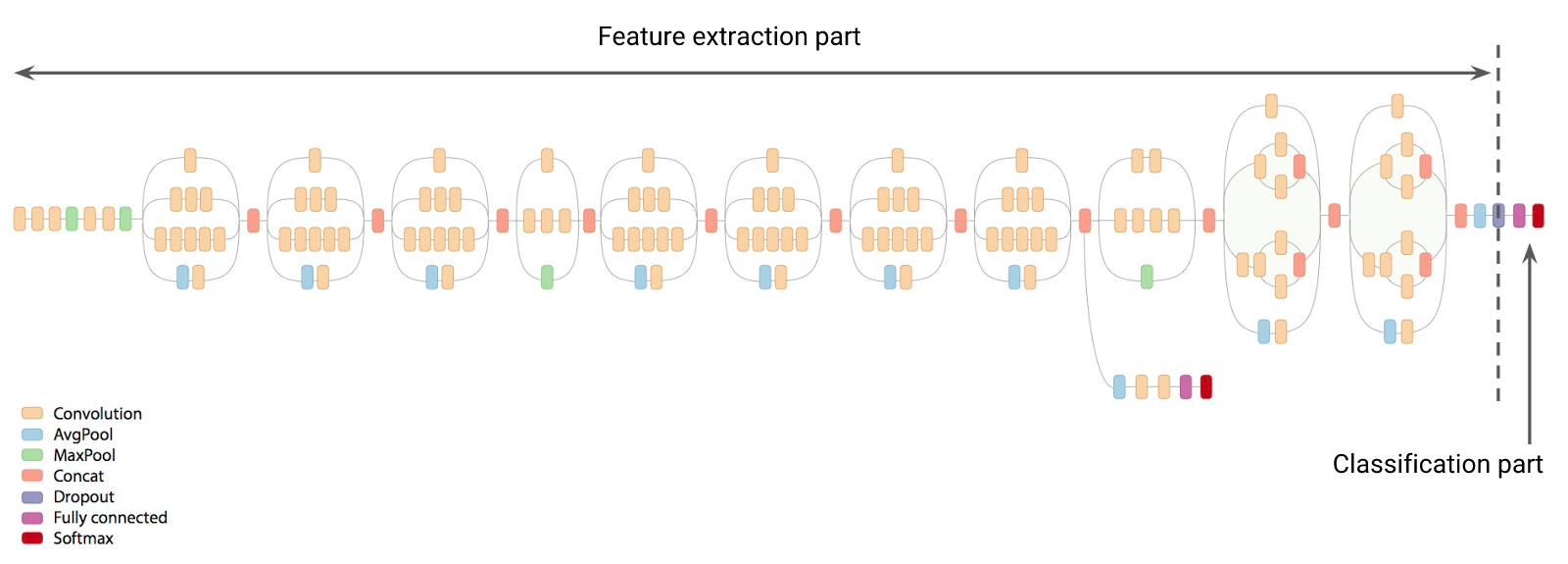}
\begin{center}Source: Adapted from~\cite[Figure 5]{devikartransfer2016}\end{center}
\end{figure}

Transfer learning is a machine learning method which utilizes a pre-trained neural network, this technique allows the detachment of the lasts outer layer (classification layer) and uses the remains structure to retraining and get new weights corresponding the classes of interest – damaged kernel in our case (Figure ~\ref{fig:featureExtraction}).

In this work, we use as a pre-trained neural network the output from~\cite{DeeperSzegedy2015}. \citet{DeeperSzegedy2015} developed a 22 layer deep convolution neural network on top of ImageNet for classifying 1000 leaf-node categories, using 1.2 million images for training, 50,000 for validation and 100,000 images for testing. 

In brief, transfer learning makes it possible to classify new classes based on a new set of images, reusing the feature extraction part and re-train the classification part of the new picture set. Since feature extraction part of the network it was already trained (which is the most complex part), the new neural network could be trained with less computational resources and time.
\subsection{Machine Learning Framework}
Nowadays we have some options to build and analyze deep neural networks using machine learning algorithms. The final choice would base on the computational infrastructure available to run this task, either the number of classes, the intended purpose and where the final Net will be deployed to handle.

For this work, it is necessary a framework for machine learning that could run on a distributed system with both CPU and GPU, with the possibility to deploy the final network to servers, desktops, mobile applications and embedded systems in an easy way.

Alongside these needs, it is necessary that the chosen framework could easily implement the Transfer Learning techniques, described before. Based on these requirements, the natural choice is TensorFlow \cite{abadi2016tensorflow}.

\citet{abadi2016tensorflow} presented TensorFlow as an interface for expressing and executing machine learning algorithms that can be performed with little or no change in a broad range of heterogeneous systems, ranging from mobile devices such as phones and tablets up to large-scale distributed systems of hundreds of machines and thousands of computational devices such as GPU cards.

\subsection{Image Pre-processing}
To generate the normal kernel images was used the FFMPEG library\footnote{FFMPEG library, see more information on \url{http://www.ffmpeg.org}.} to split the 6:25 minutes movie into 11,555 individual files (1280x720). Both normal and damaged kernel images were arranged in the separate folder for later use.

Before using the images for generating the neural network, they as randomly allocated in two distinct image sets: 80\% and 20\% for training and validation set, respectively. In the process of training the Net, other parts of wheat plant structure like spikes and leaves were used too in the composition of the DNN intending to classify better the wheat damaged kernels.

\section{Results and Rationale}
Comparing the effort described in \cite{DeeperSzegedy2015} to training a whole DNN from scratch with the number of pictures necessary to get reasonable results. In our case, the number of pictures available at the moment ($\approx$ 12,000) probably will not archive this scores and definitely, the time and computational infrastructure necessary to training and evaluate the resultant DNN must be larger than installed capacity.

A broad range of applications are using transfer learning, \citet{devikartransfer2016} describe the use in image classification of various dog breeds with an overall accuracy of 96\% from 11 dog breeds. \citet{wang2017transferring} describes the application for remote scene classification and attempt to form a baseline for transferring pre-trained DNN to remote sensing images with various spatial and spectral information.

\citet{esteva2017dermatologist} describes the use of DNN for dermatologist-level classification of skin cancer, trained end-to-end from images directly, using only pixels and disease labels as inputs for a dataset of 129,450 clinical images. 

\citet{ANNTkac2016} presents many business applications, using artificial neural networks, related to financial distress and bankruptcy problems, stock price forecasting, and decision support, with particular attention to classification tasks. For other uses of CNN and DNN, see \cite{DNNSchmidhuber2015,DeepVisualGuo2016}.

\begin{SCfigure}
  \centering
  {\caption{An illustration of Deep Convolutional Neural Network, trained using TensorFlow \cite{abadi2016tensorflow} and Transfer Learning techniques from a pre-trained Inception-v3 \cite{DeeperSzegedy2015} for an image dataset containing Fursarim Damaged Kernel, normal kernel, spikes and leaves from wheat. The output deep neural network archive an overall average accuracy of 94.7\%.}\label{fig:retrained}}
  {\includegraphics[width=0.32\textwidth]{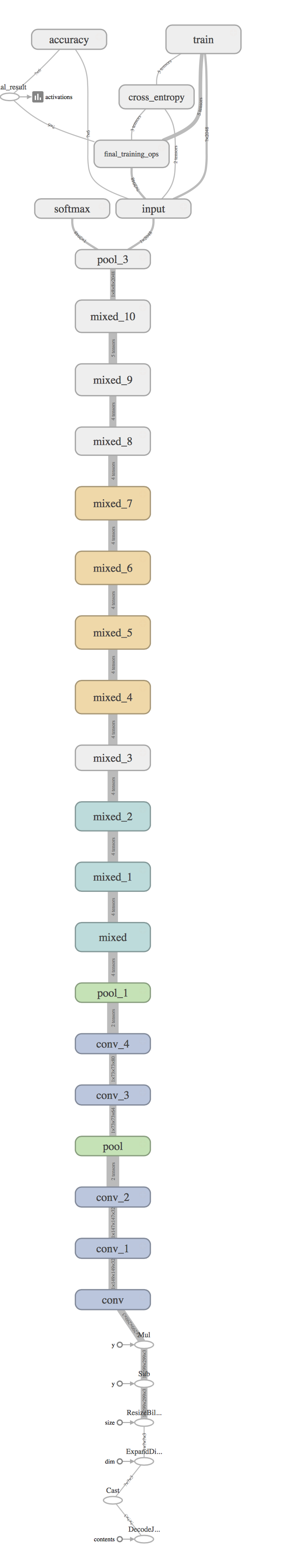}}
\end{SCfigure}

The training procedure was carried out on the described workstation and took up to 1 hour to finish, and the output retrained neural network archive an overall average accuracy of 94.7\%. The main structure of the final neural network is presented in Figure ~\ref{fig:retrained}.

For validation purposes and to check the correctness in classifying new images of wheat kernels (with and without Fusarium damage) we choose a new dataset to validate against DNN. These images were from National Research Centre for Wheat articles published over last decade about this content, alongside with other open source images of Fusarium damaged kernel found over the Internet.

At this point, we could share our positive experience in using the transfer learning techniques in relation to time to training a DNN with a new set of images and classes and the overall accuracy achieved with this initial image dataset.

The results from validation dataset present a total of 20\% of misclassification, and in half (10\%) the new images were classified as damaged leaves class. This results could be related to two main reasons: (a) the small number of pictures with Fusarium Damaged Kernels and (b) the prevalence of the normal wheat kernels present in the initial dataset ($\approx$ 80\%).
\section{Conclusions}
The associated use of Transfer Learning, TensorFlow, and Inception-v3 cut the time necessary to training and the necessity to have a large image dataset ($\approx$ 120,000) to start the classification procedure with a good accuracy level, compared to training a DNN from scratch.

The misclassification for damaged leaves class could be associated with the characteristics of damage both in kernel and leaves (in the most cases) where the region color of lesions was more blight that the standard wheat kernel.

Unfortunately, the symptoms on leaves present in this initial dataset was not separated by diseases.Thus, it was not possible to claim that some visual characteristics of FHB in leaves could be transferred to kernel evaluations in our context. This hypothesys needs to be investigated by adding new images to the present dataset related to Fusarium Damaged Kernel and wheat leaves with FHB symptoms.

Beside this misclassification for a new dataset, the overall average accuracy archived (94.7\%) for this Fusarium Damaged Kernels Deep Neural Network (FDK-DNN). Therefore, there is a potential of using this methodology for classifying Fusarium Damaged Kernel by means of smartphone camera.

The accuracy of the proposed methodology is equivalent to ones using HSI methodology  presented by~\cite{barbedo2017deoxynivalenol}.

An interesting future work could be related to using a mixed of RGB pictures, and layers from HSI operational spectra for Fusarium Damaged Kernel proposed by \cite{barbedo2015detecting}.

\bibliographystyle{abbrvnat}
\bibliography{base-referencias}

\end{document}